\pdfoutput=1
\documentclass[letterpaper]{article}
\usepackage[preprint]{aaai2027}

\usepackage{times}
\usepackage{booktabs}
\usepackage{multirow}
\usepackage{helvet}
\usepackage{courier}
\usepackage[hyphens]{url}
\usepackage{graphicx}
\usepackage{natbib}
\usepackage{caption}

\newcommand{\best}[1]{\textbf{#1}}
\newcommand{\second}[1]{\underline{#1}}
\usepackage{amsmath}
\usepackage{amssymb}

\usepackage{algorithm}
\usepackage{algorithmic}

\title{StreamSplat: Streaming Feed-Forward 3D Gaussian Splatting}

\author{
    Changhao Song,
    Yuxuan Wang,
    Qibiao Li,
    Youcheng Cai\corresponding,
    Ligang Liu
}

\affiliations{
    University of Science and Technology of China
}

\begin{document}

\maketitle

\begin{abstract}
Feed-forward 3D Gaussian Splatting enables efficient novel-view synthesis without per-scene optimization, but most existing methods assume a fixed set of context views and process them jointly. This limits their applicability to online scenarios where calibrated views arrive sequentially and the scene must be updated causally. We present \emph{StreamSplat}, a streaming feed-forward 3DGS framework that incrementally maintains a persistent geometry-grounded scene state and decodes it into renderable 3D Gaussians after each input chunk. StreamSplat centers on a \textbf{Voxel-Aligned Causal Cache (VACC)}, which stores historical 3D tokens in a memory-bounded voxel structure so that memory grows with explored scene geometry rather than stream length. To better reuse history during causal prediction, we introduce \textbf{History-Projected Depth Anchoring (HPDA)} to project cached geometry as depth guidance for current cost-volume estimation, and \textbf{Cache-Guided Feature Injection (CGFI)} to inject cached latent evidence into Gaussian-token regression. Experiments on DL3DV, RealEstate10K, and ScanNet show that StreamSplat remains competitive with state-of-the-art feed-forward 3DGS methods under sparse causal inputs, despite not using future views or full-scene context. More importantly, it scales to long input streams with 256, 512, and 1024 views where fixed-view baselines run out of memory, yielding sustained improvements in novel-view synthesis quality as more observations arrive. The code will be made publicly available upon acceptance.
\end{abstract}
\section{Introduction}
\label{sec:introduction}

Novel-view synthesis (NVS) aims to reconstruct a 3D scene from observed images and render photorealistic views from novel viewpoints. Recently, 3D Gaussian Splatting (3DGS) has made explicit radiance-field reconstruction substantially more efficient by representing scenes with anisotropic Gaussian primitives and differentiable rasterization~\cite{kerbl2023gaussians}. Owing to its high rendering quality and real-time rasterization, 3DGS has become an increasingly attractive representation for AR/VR interaction, robotic perception, autonomous navigation, and other applications that require spatially consistent visual understanding.

\begin{figure}[t!]
\centering
\includegraphics[width=\columnwidth]{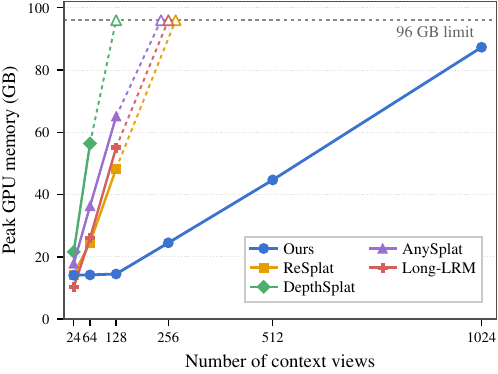}
\caption{
Peak GPU memory under increasing context lengths on ScanNet, using the
same fixed-view baselines, resolution, and context sampling as Table~\ref{tab:1},
on one NVIDIA RTX PRO 6000 (96\,GB). Baseline memory grows rapidly with
the number of jointly processed views, whereas our chunk-wise processing
and memory-bounded voxel cache stay far lower and support much longer
streams.
}
\label{fig:memory_curve}
\end{figure}

Despite its rendering efficiency, standard 3DGS still requires per-scene optimization to fit Gaussian parameters for each new scene. Feed-forward 3DGS methods~\cite{charatan2024pixelsplat,chen2024mvsplat,xu2025depthsplat,ye2025noposplat,jiang2025anysplat,li2026tokensplat} seek to remove this optimization by learning to directly predict Gaussian scene representations from input images. For example, AnySplat~\cite{jiang2025anysplat} reconstructs 3D Gaussians from unconstrained views in a feed-forward manner, extending Gaussian prediction beyond fixed and posed sparse-view settings. TokenSplat~\cite{li2026tokensplat} targets pose-free feed-forward reconstruction by aligning cross-view information in token space and jointly estimating camera poses and 3D Gaussians.

Despite these advances, existing feed-forward 3DGS methods do not directly address streaming NVS, where perceiving and reconstructing 3D geometry from videos naturally calls for online processing in interactive and low-latency applications~\cite{zhuo2025streaming4d}. In this setting, the model should produce a renderable scene estimate as new observations arrive, rather than waiting until all observations are available. This raises a natural question: how can a feed-forward 3DGS model causally process a view stream and provide anytime reconstruction as each input chunk arrives? This streaming formulation raises two key challenges: (1) the model needs a persistent memory that is spatially meaningful rather than merely temporal; and (2) incremental reconstruction must avoid the accumulation of local errors, since streaming inputs with limited overlap, occlusion, or weak texture can introduce incorrect geometry that later affects the reconstruction.

We address these challenges with \emph{StreamSplat}, a geometry-grounded memory framework for streaming feed-forward 3D Gaussian Splatting. Given a calibrated image stream, \emph{StreamSplat} processes incoming frames in chunks and builds the scene through three modules: \textbf{Voxel-Aligned Causal Cache (VACC)}, which stores historical tokens with world-space positions, latent features, appearance anchors, and confidence scores, and updates them through memory-bounded voxel fusion so that the scene state grows with explored geometry rather than the number of frames; \textbf{History-Projected Depth Anchoring (HPDA)}, which projects cached 3D tokens into the current views and converts reliable historical geometry into depth anchors for the multi-view cost volume; and \textbf{Cache-Guided Feature Injection (CGFI)}, which injects projected cache features into current Gaussian-token regression, especially when local matching is uncertain. The resulting cache can be decoded into explicit 3D Gaussians after any chunk, enabling anytime reconstruction during streaming input. We evaluate our method under a long-context streaming NVS protocol on DL3DV, RealEstate10K, and ScanNet, where it remains competitive under short contexts and shows increasing advantages as the input stream becomes longer, while fixed-view feed-forward baselines become memory-limited or fail to scale effectively.

Our contributions are summarized as follows:
\begin{itemize}
\item We propose \emph{StreamSplat}, a streaming feed-forward 3DGS framework that causally processes incoming views, updates a persistent scene state, and supports anytime Gaussian reconstruction without reprocessing the full image history.
\item We introduce \textbf{Voxel-Aligned Causal Cache (VACC)}, a 3D history memory that stores geometry-grounded historical tokens and updates them through memory-bounded voxel fusion, allowing the scene state to scale with explored geometry rather than the number of frames.
\item We design \textbf{History-Projected Depth Anchoring (HPDA)} and \textbf{Cache-Guided Feature Injection (CGFI)}, which project historical 3D tokens into the current views to guide depth estimation and condition Gaussian-token regression, reducing ambiguity and error accumulation in streaming reconstruction.
\end{itemize}
\section{Related Work}
\paragraph{Optimization-based 3D Gaussian Splatting.}
NeRF~\cite{mildenhall2021nerf} demonstrated high-quality novel-view synthesis with neural radiance fields, but its volumetric rendering and per-scene optimization are computationally expensive. 3D Gaussian Splatting (3DGS)~\cite{kerbl2023gaussians} replaces implicit fields with explicit anisotropic Gaussian primitives and enables real-time differentiable rasterization. Subsequent methods improve rendering quality, compactness, and optimization efficiency, including Mip-Splatting~\cite{yu2024mip}, Mini-Splatting~\cite{fang2024mini}, EDGS~\cite{kotovenko2026edgs}, Plug-and-Play PDE Optimization~\cite{Mo_2026_CVPR}, FastGS~\cite{ren2026fastgs}, and Faster-GS~\cite{hahlbohm2026faster}. These methods substantially advance per-scene Gaussian optimization, while our work focuses on feed-forward streaming reconstruction.

\paragraph{Feed-forward 3D Gaussian Splatting.}
Feed-forward 3DGS methods amortize scene reconstruction across training data and directly predict Gaussian representations at inference time. pixelSplat~\cite{charatan2024pixelsplat} predicts Gaussians from image pairs, and MVSplat~\cite{chen2024mvsplat} introduces plane-sweep cost volumes for explicit cross-view matching. DepthSplat~\cite{xu2025depthsplat} further incorporates monocular depth cues to improve robustness. Several methods relax the need for calibrated inputs: NoPoSplat~\cite{ye2025noposplat} predicts Gaussians in a canonical camera frame, Splatt3R~\cite{smart2024splatt3r} builds on learned stereo geometry, and AnySplat~\cite{jiang2025anysplat} jointly reasons about cameras and Gaussians from unconstrained views. ReSplat~\cite{xu2025resplat} recurrently refines compact Gaussian representations, while TokenSplat~\cite{li2026tokensplat} and ZipSplat~\cite{veicht2026zipsplat} decouple the primitive budget from image pixels. These works establish efficient feed-forward Gaussian prediction, whereas our setting requires online scene updates from a sequential view stream.

\paragraph{Streaming/Online 3D Gaussian Splatting.}
Recent systems maintain Gaussian scenes from sequential observations. StreamGS~\cite{li2025streamgs} reconstructs per-frame Gaussians and merges redundant primitives for online unposed reconstruction. StreamSplat~\cite{wu2026streamsplat} extends streaming Gaussian reconstruction to dynamic scenes with persistent Gaussian propagation and adaptive fusion. LongSplat~\cite{huang2026longsplat} compresses historical Gaussians over long sequences, and OF$^3$GS~\cite{chen2026of3gs} performs on-the-fly feed-forward reconstruction from unposed images, incrementally registering each incoming view into a growing Gaussian scene. ReCoSplat~\cite{cheng2026recosplat} autoregressively predicts Gaussians from posed or unposed streams and compresses its attention KV cache to keep long sequences tractable. These methods motivate streaming 3DGS, while our work emphasizes a geometry-grounded spatial cache that can be queried by subsequent feed-forward prediction.

\begin{figure*}[t]
    \centering
    \includegraphics[width=\textwidth]{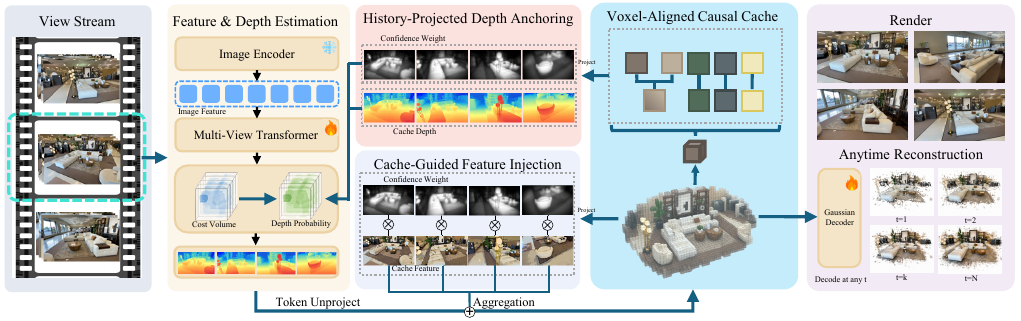}
    \caption{
    Overview of \emph{StreamSplat}. The view stream is processed chunk by
    chunk: \textbf{VACC} maintains a geometry-grounded 3D history memory,
    \textbf{HPDA} projects cached geometry into the current views as depth
    anchors, and \textbf{CGFI} reuses cached features to condition
    Gaussian-token regression.
    }
    \label{fig:pipeline}
\end{figure*}

\section{Method}
\label{sec:method}

\subsection{Problem Formulation}
\label{sec:problem_formulation}

Given a calibrated view stream $\mathcal{V}=\{(I_t,C_t)\}_{t=1}^{N}$, where $N$ is the number of context views, $I_t$ is the image observed at streaming step $t$, and $C_t$ denotes its camera parameters, our goal is to reconstruct a 3D Gaussian scene in a streaming feed-forward manner. Unlike fixed-view reconstruction, where all context views are available before inference, streaming reconstruction observes the scene progressively and must update its scene estimate as new observations arrive. At step $t$, a network $F_\theta$ consumes the current observation $(I_t,C_t)$ and the historical token cache $\mathcal{H}_{t-1}$, and outputs newly predicted tokens $\mathcal{T}_t$, the updated cache $\mathcal{H}_t$, and the current Gaussian scene $\mathcal{G}_t$:
\begin{equation}
\left(\mathcal{T}_t,\mathcal{H}_t,\mathcal{G}_t\right)
=
F_\theta\left(I_t,C_t,\mathcal{H}_{t-1}\right),
\label{eq:streaming_problem}
\end{equation}
where $\theta$ denotes the learnable parameters of the feed-forward model. This formulation gives the model an anytime reconstruction property: online applications may decode the current scene immediately after each chunk, while offline evaluation can decode only after the final context view.

\subsection{Overview}
\label{sec:method_overview}

As illustrated in Fig.~\ref{fig:pipeline}, \emph{StreamSplat} consists of a geometry-grounded cache and two cache-guided prediction modules. First, \textbf{Voxel-Aligned Causal Cache (VACC)} maintains the 3D history memory, where tokens are anchored in a common world coordinate system and fused through a memory-bounded voxel update. Second, \textbf{History-Projected Depth Anchoring (HPDA)} projects the previous cache into the current views and uses reliable cached geometry as depth anchors for the multi-view cost volume. Third, \textbf{Cache-Guided Feature Injection (CGFI)} injects projected cache features into the token-regression input so that historical evidence can condition current Gaussian prediction. Together, HPDA and CGFI make the feed-forward predictor aware of the persistent 3D history, while VACC writes newly predicted tokens back into the scene state.

\subsection{Base framework.}
Our method is built on a feed-forward Gaussian reconstruction pipeline following ReSplat~\cite{xu2025resplat}, where compact 3D tokens are decoded by a 3D token mixer and a Gaussian prediction head into explicit 3D Gaussians. We use DepthSplat~\cite{xu2025depthsplat} as the image encoder and geometry prediction backbone. Given calibrated context views $\{I_i,C_i\}_{i=1}^{N}$, the backbone converts the images and cameras into 3D tokens:
\begin{equation}
\{I_i,C_i\}_{i=1}^{N}
\rightarrow
\{(p_j,f_j)\}_{j=1}^{M},
\label{eq:base_token_prediction}
\end{equation}
where $p_j$ is a world-space token position and $f_j$ is its latent feature. With a $4\times$ downsampled feature grid, the token number is $M=N H W/16$.

Specifically, token positions are obtained from cost-volume-based depth estimation. For each reference view $i$, the backbone samples depth candidates $\{d_k\}_{k=1}^{D}$ and warps source-view features onto the reference image plane at each candidate depth using the known intrinsics and extrinsics. Dot-product correlations between the reference feature and the warped source features form a plane-sweep cost volume $C_i(k,u)$ for pixel $u$. A 2D U-Net decoder then combines the cost volume with image features and predicts a depth probability distribution $P_i(k\mid u)$ through a softmax over depth candidates. The final depth is computed by the expectation over all candidates:
\begin{equation}
\hat d_i(u)=\sum_{k=1}^{D}P_i(k\mid u)d_k.
\label{eq:base_depth_regression}
\end{equation}
The predicted depth is then back-projected with the camera parameters to obtain 3D token positions. The corresponding token feature $f_i(u)$ is predicted from the fused image feature, depth distribution, and decoder feature at the same pixel. On top of this backbone, VACC, HPDA, and CGFI turn the fixed-view feed-forward architecture into a streaming model: VACC stores historical 3D tokens, HPDA injects historical geometry into depth estimation, and CGFI reuses historical features during Gaussian-token regression.

\subsection{Voxel-Aligned Causal Cache}
\label{sec:vacc}

\textbf{Voxel-Aligned Causal Cache (VACC)} is the persistent 3D memory of \emph{StreamSplat}. At each streaming step, the base framework produces current 3D tokens $\{(p_j,f_j)\}$ from the newly observed views. VACC stores these tokens in a spatially indexed cache so that historical evidence can be reused without keeping all past images or feature maps. Motivated by the long-term spatial token cache in STAC~\cite{wang2026stac}, VACC exploits redundancy in 3D space: repeated observations of the same local surface are fused, while new reliable geometry is inserted into the cache.

For each incoming token $m_\ell = (p_\ell,f_\ell)\in\{(p_j,f_j)\}_{j=1}^{M}$, VACC first voxelizes the token position with voxel size $\Delta$ and assigns it to voxel $b_\ell$.
The update is then performed independently inside voxel $b_\ell$. Each voxel stores at most $K$ pivot tokens, where each pivot summarizes a local appearance mode using an aggregated token state, a confidence weight, and a high-confidence seed feature for future matching. The confidence weight $\omega_\ell$ is taken as the peak of the depth probability distribution in Eq.~\eqref{eq:base_depth_regression}, i.e., $\omega_\ell=\max_k P_i(k\mid u)$, and low-confidence incoming tokens with $\omega_\ell<\tau_{\mathrm{conf}}$ are ignored to avoid polluting the long-term cache, where we set $\tau_{\mathrm{conf}}=0.3$ in all experiments.

Let $\mathcal{H}_{b_\ell}=\{m^*_r\}_{r=1}^{n_{b_\ell}}$ denote the pivots currently stored in voxel $b_\ell$, with $n_{b_\ell}\le K$. If the voxel is not full, VACC directly inserts the incoming token as a new pivot. If the voxel already contains $K$ pivots, VACC temporarily forms a candidate set by adding the incoming token and then merges the most similar pair:
\begin{equation}
\mathcal{H}_{b_\ell}' =
\begin{cases}
\mathcal{H}_{b_\ell}\cup\{m_\ell\}, & n_{b_\ell}<K,\\
\operatorname{MergeClosest}(\mathcal{H}_{b_\ell}\cup\{m_\ell\}), & n_{b_\ell}=K.
\end{cases}
\label{eq:vacc_capacity_update}
\end{equation}
where $\operatorname{MergeClosest}$ selects the two tokens $m_0$ and $m_1$ with the highest cosine similarity between their features and replaces them with a merged token $m_\star$.

For the selected pair $m_0$ and $m_1$, the merged token $m_\star$ is computed by confidence-weighted aggregation:
\begin{equation}
\begin{aligned}
p_\star &= (\omega_0p_0+\omega_1p_1)/(\omega_0+\omega_1),\\
f_\star &= (\omega_0f_0+\omega_1f_1)/(\omega_0+\omega_1),\\
\omega_\star &= \max(\omega_0,\omega_1).\\
\end{aligned}
\label{eq:vacc_pair_selection_and_merge}
\end{equation}
The merged pivot replaces the selected pair, so the voxel again contains exactly $K$ pivots. We keep the seed feature from the higher-confidence member of the merged pair, which preserves a reliable representative for future similarity matching.

In our experiments, we set $K=4$ and $\Delta=0.04$ in the world units of the input camera poses. Since the voxel size is sufficiently small, tokens assigned to the same voxel are spatially close and usually correspond to the same local region, making the within-voxel merge preserve fine geometric detail while removing redundant observations. Because VACC stores at most $K$ pivots in each occupied voxel, the memory grows with explored scene volume and voxel resolution, rather than with stream length, and provides the historical geometry and latent features used by HPDA and CGFI. 

\subsection{History-Projected Depth Anchoring}
\label{sec:hpda}

\textbf{History-Projected Depth Anchoring (HPDA)} uses the cached 3D history as geometric guidance for the current views. The previous cache $\mathcal{H}_{t-1}$ is projected into each current view:
\begin{equation}
\left(
D_{\mathcal{H}}^v,
\Omega_{\mathcal{H}}^v,
F_{\mathcal{H}}^v
\right) 
=
\mathcal{P}(\mathcal{H}_{t-1}, C_v),
\label{eq:cache_projection}
\end{equation}
where $C_v$ is the camera parameter of view $v$, and $D_{\mathcal{H}}^v$, $\Omega_{\mathcal{H}}^v$, and $F_{\mathcal{H}}^v$ denote the projected cache depth, confidence weight, and confidence-weighted feature maps.

Let $\{d_k\}_{k=1}^{D}$ be the depth candidates of the multi-view cost volume. For valid cache pixels, HPDA converts the projected cache depth into a confidence-aware Gaussian anchor over candidate depths, with mean $\mu_{\mathcal{H}}^v(u)=D_{\mathcal{H}}^v(u)$ and variance $[\sigma_{\mathcal{H}}^v(u)]^2$, where $\sigma_{\mathcal{H}}^v(u)=\sigma_0+\lambda_{\sigma}(1-\Omega_{\mathcal{H}}^v(u))$:
\begin{equation}
\begin{aligned}
P_{\mathcal{H}}^v(k,u)
&=
\Omega_{\mathcal{H}}^v(u)
\exp\left(
-\frac{
[d_k-\mu_{\mathcal{H}}^v(u)]^2
}{
2[\sigma_{\mathcal{H}}^v(u)]^2
}
\right).
\end{aligned}
\label{eq:cache_depth_prior}
\end{equation}
High-confidence cached geometry produces a sharper anchor, while uncertain cached observations produce a broader one. Given the depth probability $P_i(k\mid u)$ from Eq.~\eqref{eq:base_depth_regression}, HPDA directly fuses it with the cache anchor and renormalizes the result:
\begin{equation}
\begin{aligned}
P^v(k\mid u)
&=
\frac{
P_i(k\mid u)+\gamma_d P_{\mathcal{H}}^v(k,u)
}{
\sum_{k'=1}^{D}
\left[
P_i(k'\mid u)+\gamma_d P_{\mathcal{H}}^v(k',u)
\right]
}.
\end{aligned}
\label{eq:cache_prior_injection}
\end{equation}
Here $P^v(k\mid u)$ is the cache-anchored depth probability distribution. The base anchor variance and the confidence-dependent variance scaling are fixed to $\sigma_0=0.1$ and $\lambda_{\sigma}=0.5$, while the cache-anchor strength $\gamma_d$ is a learnable scalar optimized jointly with the network rather than a manually tuned hyperparameter, so that the model itself decides how much to rely on cached geometry. HPDA therefore lets the current views reuse reliable historical geometry while still allowing the cost volume to revise uncertain cached structure.

\subsection{Cache-Guided Feature Injection}
\label{sec:cgfi}

\textbf{Cache-Guided Feature Injection (CGFI)} complements HPDA by reusing historical appearance and latent evidence during Gaussian-token regression. The projected cache feature map $F_{\mathcal{H}}^v$ contains feature evidence accumulated from previous chunks. Following the confidence definition in VACC, let $\omega^v(u)=\max_k P^v(k\mid u)$ denote the current confidence, i.e., the peak of the cache-anchored depth distribution. Since pixels with high current confidence already have reliable local evidence, we attenuate the cache feature by the current uncertainty:
\begin{equation}
\widehat{F}_{\mathcal{H}}^v(u)
=
\left(1-\omega^v(u)\right)F_{\mathcal{H}}^v(u).
\label{eq:cache_feature_attenuation}
\end{equation}
The attenuated cache feature is concatenated with the current token feature and aggregated by a 2D U-Net:
\begin{equation}
\widetilde f^v(u)
=
\Phi_{\mathrm{agg}}
\left(
\operatorname{Concat}
\left(
f^v(u),\,
\widehat{F}_{\mathcal{H}}^v(u)
\right)
\right).
\label{eq:token_regressor_input}
\end{equation}
Here $\Phi_{\mathrm{agg}}$ is a 2D U-Net aggregation network, $f^v(u)$ is the current latent token feature before CGFI, and $\widetilde f^v(u)$ is the cache-enhanced feature used for Gaussian-token regression. The token position is obtained by back-projecting the predicted depth. In this way, CGFI injects historical scene evidence only where it is useful, helping the model reduce local ambiguity without introducing an additional temporal feature memory. The predicted tokens are then passed to VACC for causal cache update, closing the streaming reconstruction loop.

\begin{figure*}[t!]
\centering
\includegraphics[width=\textwidth]{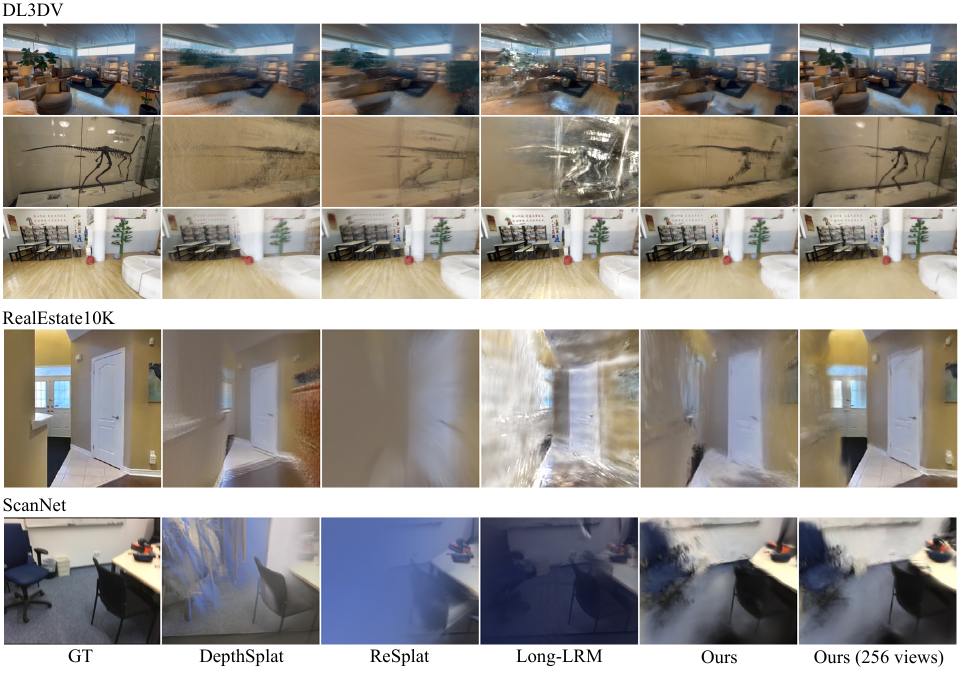}
\caption{
Qualitative comparisons against DepthSplat, ReSplat, and Long-LRM on
DL3DV (top three rows), RealEstate10K (fourth row), and ScanNet (last
row). All methods use \textbf{64 context views}, except the last column,
which uses 256 views and shows the additional gain our streaming design
obtains from a longer input stream.
}
\label{fig:qualitative}
\end{figure*}

\begin{table*}[t!]
\centering
\caption{
Quantitative results on DL3DV, RealEstate10K, and ScanNet. ``Mem.'' is peak
GPU memory in GB and ``OOM'' denotes out-of-memory. \best{Bold} and \second{underline}
mark the best and second-best per dataset and context length.
}
\label{tab:1}

\footnotesize
\setlength{\tabcolsep}{2.4pt}
\renewcommand{\arraystretch}{1.08}

\begin{tabular*}{\textwidth}{
@{\extracolsep{\fill}}
ll
*{3}{cccc}
@{}
}
\toprule
Dataset
& Method
& \multicolumn{4}{c}{24 Views}
& \multicolumn{4}{c}{64 Views}
& \multicolumn{4}{c}{128 Views} \\

\cmidrule(lr){3-6}
\cmidrule(lr){7-10}
\cmidrule(lr){11-14}

&
& PSNR$\uparrow$ & SSIM$\uparrow$ & LPIPS$\downarrow$ & Mem.$\downarrow$
& PSNR$\uparrow$ & SSIM$\uparrow$ & LPIPS$\downarrow$ & Mem.$\downarrow$
& PSNR$\uparrow$ & SSIM$\uparrow$ & LPIPS$\downarrow$ & Mem.$\downarrow$ \\
\midrule

DL3DV
& DepthSplat
& 13.41 & 0.372 & 0.547 & 13.3
& 15.02 & 0.434 & 0.517 & 34.4
& \multicolumn{4}{c}{OOM} \\

& AnySplat
& 10.97 & 0.400 & 0.581 & 15.4
& 13.97 & 0.475 & 0.483 & 34.3
& 16.76 & 0.536 & 0.413 & 64.5 \\

& ReSplat
& \best{14.48} & \best{0.455} & \underline{0.545} & 13.4
& \best{16.80} & \underline{0.552} & \underline{0.439} & 34.1
& \best{19.10} & \best{0.642} & \best{0.346} & 67.1 \\

& Long-LRM
& 13.29 & \underline{0.448} & \best{0.534} & 15.1
& \underline{16.74} & \best{0.579} & \best{0.414} & 40.2
& 17.50 & \underline{0.611} & \underline{0.360} & 75.5 \\

& OF$^3$GS
& \underline{14.32} & 0.380 & 0.580 & \best{8.3}
& 14.45 & 0.370 & 0.584 & \best{10.7}
& 14.16 & 0.357 & 0.606 & \best{16.7} \\

& Ours
& 12.95 & 0.379 & 0.565 & \underline{11.3}
& 15.38 & 0.498 & 0.479 & \underline{25.4}
& \underline{18.36} & 0.594 & 0.414 & \underline{49.1} \\

\midrule

RealEstate10K
& DepthSplat
& 12.92 & 0.413 & 0.553 & \underline{7.9}
& 13.71 & 0.445 & 0.538 & 19.9
& 14.31 & 0.473 & 0.516 & 39.2\\

& AnySplat
& 10.97 & \best{0.500} & 0.557 & 15.3
& 13.57 & \best{0.565} & \underline{0.462} & 33.9
& 17.54 & \underline{0.662} & \underline{0.344} & 64.3\\

& ReSplat
& \best{13.62} & \underline{0.463} & \best{0.528} & 8.1
& \best{15.93} & 0.551 & 0.474 & 19.9
& \underline{18.40} & 0.644 & 0.377 & 38.8 \\

& Long-LRM
& 11.27 & 0.417 & 0.563 & 8.9
& \underline{15.85} & \underline{0.554} & \best{0.425} & 22.7
& \best{18.77} & \best{0.675} & \best{0.313} & 45.8 \\

& OF$^3$GS
& \underline{13.21} & 0.437 & 0.589 & \best{6.6}
& 13.70 & 0.445 & 0.575 & \best{8.0}
& 13.68 & 0.437 & 0.574 & \best{11.5} \\

& Ours
& 13.03 & 0.431 & \underline{0.543} & \best{6.6}
& 14.94 & 0.511 & 0.495 & \underline{14.6}
& 17.84 & 0.645 & 0.401 & \underline{28.0}\\

\midrule

ScanNet
& DepthSplat
& 11.43 & 0.432 & 0.625 & 21.6
& 11.66 & 0.450 & 0.631 & 56.4
& \multicolumn{4}{c}{OOM} \\

& AnySplat
& 7.93 & 0.468 & 0.670 & 17.8
& 9.16 & 0.493 & 0.643 & 36.3
& 10.63 & 0.516 & 0.620 & 65.1\\

& ReSplat
& \underline{13.33} & 0.498 & \best{0.558} & 14.3
& 13.64 & 0.518 & \best{0.554} & 24.6
& 14.30 & 0.544 & \best{0.545} & 48.3 \\

& Long-LRM
& 12.59 & \best{0.528} & \underline{0.587} & \best{10.3}
& \underline{14.03} & \best{0.564} & \underline{0.562} & 26.0
& \underline{14.57} & \best{0.579} & \underline{0.553} & 55.1\\

& OF$^3$GS
& 12.43 & 0.484 & 0.667 & 15.6
& 12.47 & 0.480 & 0.666 & \underline{15.8}
& 12.47 & 0.479 & 0.664 & \underline{16.2} \\

& Ours
& \best{13.71} & \underline{0.521} & 0.610 & \underline{14.1}
& \best{14.36} & \underline{0.547} & 0.602 & \best{14.2}
& \best{15.12} & \underline{0.573} & 0.592 & \best{14.5} \\

\bottomrule
\end{tabular*}
\end{table*}

\paragraph{Gaussian Decoder.}
The final scene representation is obtained by decoding cache-enhanced 3D
tokens into explicit Gaussian primitives. For a set of predicted tokens
$\{(p_j,\widetilde f_j)\}_{j=1}^{M}$, we directly use the 3D positions
$\{p_j\}_{j=1}^{M}$ as Gaussian centers. A lightweight Gaussian head, implemented as a two-layer MLP, maps each context-aggregated feature $\widetilde f_j$ to the remaining Gaussian parameters, including opacity, color coefficients, scale, and rotation. During training, the decoder is applied to the concatenated gradient-retaining tokens from all chunks for end-to-end rendering supervision. During inference, it is applied to the final fused cache, enabling the current streaming state to be converted into renderable 3D Gaussians at any time.

\subsection{Training Loss}
\label{sec:training}

Following DepthSplat~\cite{xu2025depthsplat}, we supervise the full model on target views using a rendering loss that combines mean squared error (MSE) and LPIPS:
\begin{equation}
\mathcal{L}_{\mathrm{render}}
=
\sum_{r}
\left[
\mathrm{MSE}(\hat I_r, I_r)
+
\lambda_{\mathrm{LPIPS}}\ell_{\mathrm{LPIPS}}(\hat I_r, I_r)
\right],
\label{eq:training_loss}
\end{equation}
where $r$ indexes the target views, $\hat I_r$ and $I_r$ denote the rendered and ground-truth images, and $\lambda_{\mathrm{LPIPS}}$ balances the two terms.

\section{Experiments}
\label{sec:experiments}
\subsection{Implementation Details}
\label{sec:implementation_details}
We implement our method in PyTorch and optimize \emph{StreamSplat} for 100,000 iterations using AdamW with a learning rate of $1\times10^{-4}$ and a batch size of one. All experiments are conducted on a single NVIDIA RTX PRO 6000 GPU with 96GB of memory. Our Gaussian splatting renderer is based on the Mip-Splatting implementation~\cite{yu2024mip}. During both training
and inference, the input stream is processed causally in chunks of 4 views. On
ScanNet with 64 input views, the end-to-end latency from receiving an input
chunk to rendering one novel view is about 143\,ms, i.e.\ 6.98\,FPS.

\subsection{Datasets and Metrics}
\label{sec:datasets_metrics}

StreamSplat is trained on over 10,000 scenes from DL3DV~\cite{ling2024dl3dv}, over 7,000 trajectories from RealEstate10K~\cite{zhou2018stereo}, and 100 scenes from ScanNet~\cite{dai2017scannet}. For evaluation, we report results on the official DL3DV-140 benchmark, 140 randomly sampled scenes from the RealEstate10K test split, and the 20 longest ScanNet test scenes, each containing approximately 7,000 frames on average.

To test streaming reconstruction, we reconstruct a 3D Gaussian scene using varying numbers of context views, including 24, 64, 128, 256, 512, and 1024 views. For each test scene, the context views are taken from the sequential camera trajectory to simulate an online stream. After processing the specified number of context views, we decode the current scene representation into 3D Gaussians and evaluate novel-view synthesis quality on 64 uniformly sampled test views. We report PSNR, SSIM~\cite{wang2004image}, and LPIPS~\cite{zhang2018unreasonable}, averaged over all test views and test scenes.

\subsection{Comparative Evaluation}
\label{sec:main_results}

\textbf{Baselines.}
We compare against representative feed-forward 3D Gaussian reconstruction
methods, including DepthSplat~\cite{xu2025depthsplat},
AnySplat~\cite{jiang2025anysplat}, ReSplat~\cite{xu2025resplat}, and
Long-LRM~\cite{chen2024longlrm}, together with
OF$^3$GS~\cite{chen2026of3gs}, an on-the-fly feed-forward method that also
consumes views incrementally and is evaluated with its official checkpoint.
Other recent streaming Gaussian methods~\cite{li2025streamgs,huang2026longsplat} provide no public implementation, and \citet{wu2026streamsplat} targets a different objective, so none of them can be evaluated under our protocol. When a method cannot process a given number of context views under the same hardware and resolution, we report out-of-memory as ``OOM''.

\textbf{Results.}
Table~\ref{tab:1} reports quantitative comparisons under increasing numbers of context views. With sparse inputs (24, 64, and 128 views), \emph{StreamSplat} remains comparable to existing state-of-the-art feed-forward 3DGS methods across DL3DV, RealEstate10K, and ScanNet. This is a more challenging setting for our method, since the input is causal and sequential rather than globally available, so the model cannot rely on future views or full-scene context during reconstruction. Even under this constraint, our method achieves competitive quality. OF$^3$GS attains the lowest peak memory on DL3DV and RealEstate10K, but its quality barely responds to additional views: on DL3DV its PSNR moves from 14.32 at 24 views to 14.16 at 128 views, whereas ours rises from 12.95 to 18.36 over the same range.

Table~\ref{tab:2} further evaluates whether longer context leads to better reconstruction quality on ScanNet, whose indoor sequences contain thousands of frames. As the input stream grows longer, our streaming design further supports 256, 512, and 1024 views, while all fixed-view baselines run out of memory under the same hardware and resolution. \emph{StreamSplat} consistently improves as the context increases from 256 to 512 and 1024 views: PSNR rises from 15.85 to 18.07, with corresponding gains in SSIM and LPIPS. Among the baselines, only the on-the-fly OF$^3$GS keeps running past 128 views, but its quality saturates instead of benefiting from the longer stream, staying at 12.48 PSNR with 256 views and 12.31 with 512 views before also reaching out-of-memory at 1024 views. This contrast shows that the persistent cache not only enables long-sequence processing, but also effectively accumulates additional observations into a higher-quality scene representation.

Figure~\ref{fig:qualitative} compares renderings under a common budget of 64
context views, where our method is on par with the strongest feed-forward
baselines. The rightmost column uses 256 context views: the longer stream
recovers more complete scene structure and reduces artifacts in
insufficiently observed regions.

\begin{table}[t]
    \centering
        \caption{
        Results at extended context lengths, with peak memory in GB.
        OF$^3$GS is the only baseline that runs beyond 128 views.
        \best{Bold} marks the better value in each pair.
        }
    \label{tab:2}
    \footnotesize
    \setlength{\tabcolsep}{2.2pt}
    \begin{tabular*}{\columnwidth}{@{\extracolsep{\fill}}lllcccc@{}}
      \toprule
      Dataset & Views & Method
      & PSNR$\uparrow$ & SSIM$\uparrow$ & LPIPS$\downarrow$ & Mem.$\downarrow$ \\
      \midrule

      \multirow{2}{*}{DL3DV} & \multirow{2}{*}{256}
        & OF$^3$GS & 14.00 & 0.355 & 0.618 & \best{28.6} \\
      & & Ours & \best{20.15} & \best{0.637} & \best{0.392} & 93.1 \\

      \midrule
      \multirow{2}{*}{RealEstate10K} & \multirow{2}{*}{256}
        & OF$^3$GS & 13.44 & 0.430 & 0.585 & \best{18.5} \\
      & & Ours & \best{19.18} & \best{0.669} & \best{0.379} & 53.1 \\

      \midrule
      \multirow{6}{*}{ScanNet} & \multirow{2}{*}{256}
        & OF$^3$GS & 12.48 & 0.483 & 0.659 & \best{23.1} \\
      & & Ours & \best{15.85} & \best{0.597} & \best{0.582} & 24.5 \\
      \cmidrule(l){2-7}
      & \multirow{2}{*}{512}
        & OF$^3$GS & 12.31 & 0.482 & 0.667 & \best{41.4} \\
      & & Ours & \best{16.98} & \best{0.637} & \best{0.568} & 44.7 \\
      \cmidrule(l){2-7}
      & \multirow{2}{*}{1024}
        & OF$^3$GS & \multicolumn{3}{c}{--} & OOM \\
      & & Ours & \best{18.07} & \best{0.669} & \best{0.549} & 87.3 \\

      \bottomrule
    \end{tabular*}%
  
\end{table}

\subsection{Scalability and Memory Analysis}
\label{sec:memory_scaling}

As shown in Figure~\ref{fig:memory_curve}, existing fixed-view feed-forward
methods process all context views jointly, so peak memory grows rapidly and
soon reaches the hardware limit. \emph{StreamSplat} instead processes the
stream chunk by chunk and compresses history into a voxel cache, so memory is
governed by the current chunk and the retained scene representation rather than
all previously observed views. This is what allows it to keep incorporating new
observations at the long context lengths reported in Table~\ref{tab:2}.

\subsection{Ablation Studies}
\label{sec:ablation}

\begin{figure}[t]
\centering
\includegraphics[width=\columnwidth]{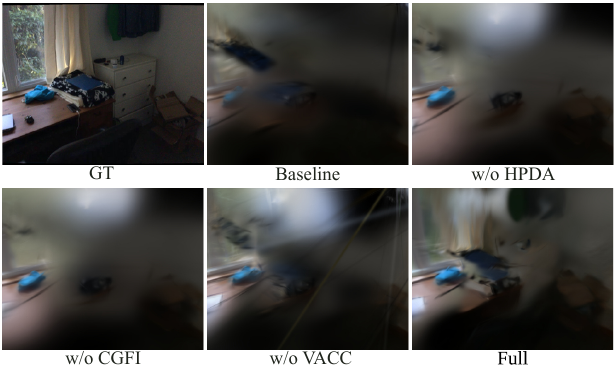}
\caption{
Qualitative ablation study on ScanNet with 256 context views.
}
\label{fig:ablation}
\end{figure}

\begin{table}[t]
\centering
\caption{
Ablation studies on ScanNet. All variants use 256 context views
unless noted.
}
\label{tab:ablation}

\footnotesize
\setlength{\tabcolsep}{2.5pt}
\renewcommand{\arraystretch}{1.08}

\begin{tabular}{lcccc}
\toprule
Variant
& PSNR$\uparrow$
& SSIM$\uparrow$
& LPIPS$\downarrow$
& Mem. (GB)$\downarrow$ \\
\midrule
Baseline
& 15.57 & 0.591 & 0.583 & 43.4 \\
w/o VACC
& 16.14 & 0.606 & 0.570 & 41.3 \\

w/o VACC (512 views)
& \multicolumn{3}{c}{--} & OOM\\
w/o HPDA
& 14.61 & 0.553 & 0.613 & 39.3 \\
w/o CGFI
& 14.77 & 0.556 & 0.612 & 28.0 \\
Full
& 15.85 & 0.597 & 0.582 & 24.5 \\
Full (512 views)
& 16.98 & 0.637 & 0.568 & 44.7 \\

\bottomrule
\end{tabular}
\end{table}

We ablate the contribution of VACC, HPDA, and CGFI on ScanNet in Table~\ref{tab:ablation} and Figure~\ref{fig:ablation}. The \emph{Baseline} disables all three modules, using an append-all cache without HPDA or CGFI; \emph{w/o VACC} keeps HPDA and CGFI but replaces voxel merging with the append-all cache; and \emph{w/o HPDA} or \emph{w/o CGFI} removes the corresponding guidance module from the full model. The modules act along different axes: VACC trades quality for memory, while HPDA and CGFI improve quality and also reduce memory. These two effects partially cancel in the \emph{Baseline} row, which is therefore not a lower bound for the single-module ablations.

\textbf{Effect of VACC.}
Replacing VACC with an append-all cache slightly improves the 256-view score but greatly increases peak memory from 24.5\,GB to 41.3\,GB. VACC is thus an explicit accuracy--memory trade-off: by merging redundant historical tokens into voxel-aligned pivots, it costs 0.29\,dB at 256 views but enables the full model to scale to 512 views and reach 16.98 PSNR, whereas the append-all cache runs out of memory.

\textbf{Effect of HPDA.}
Removing HPDA causes a clear quality drop, reducing PSNR from 15.85 to 14.61 and increasing LPIPS from 0.582 to 0.613, while peak memory rises to 39.3\,GB. This indicates that projecting cached geometry into the current cost volume provides useful depth guidance across chunks. The qualitative results in Figure~\ref{fig:ablation} also show less stable structure without HPDA.

\textbf{Effect of CGFI.}
Removing CGFI similarly degrades reconstruction quality, lowering PSNR to 14.77 and increasing LPIPS to 0.612, while peak memory rises to 28.0\,GB. This confirms that cached latent features complement the geometric guidance from HPDA during Gaussian-token regression, helping the model recover sharper and more complete novel views. Both modules also lower peak memory because they keep per-chunk predictions consistent with the cache, so repeated observations of the same surface fall into the same voxels and are absorbed by the within-voxel merge of Eq.~\eqref{eq:vacc_capacity_update} instead of occupying new ones.

\section{Conclusion}

We presented \emph{StreamSplat}, a streaming feed-forward 3DGS framework for causal novel-view synthesis from continuously arriving calibrated views. Instead of processing all context views jointly, it maintains a persistent scene state in VACC, a memory-bounded voxel cache of historical 3D tokens, and reuses that history through HPDA and CGFI to guide depth estimation and Gaussian-token regression, so a renderable scene can be decoded after each input chunk. On DL3DV, RealEstate10K, and ScanNet, StreamSplat stays competitive under sparse causal inputs and, more importantly, scales to long input streams where fixed-view baselines run out of memory, with quality improving as more views arrive.


\clearpage
\appendix
\setcounter{secnumdepth}{2}

\section{Implementation Details}
\label{sec:supp_impl}

\subsection{Computing Infrastructure and Randomness}
\label{sec:supp_infra}

All training and evaluation runs use a single NVIDIA RTX PRO 6000 Blackwell
Server Edition GPU with 96\,GB of memory, running Ubuntu 22.04.5 LTS with
NVIDIA driver 580.95 and CUDA 12.8.

Sampling randomness enters only during training, through scene and view
sampling. A single global seed is set per process, offset by the process rank,
and the train, validation and test dataloaders each carry their own fixed seed.
Evaluation draws its context and target views from a fixed precomputed index.
We do not enforce deterministic CUDA kernels, so repeated runs may differ in
the last reported digit.

\section{Evaluation Protocol}
\label{sec:supp_protocol}

\subsection{Context and Target Construction}
\label{sec:supp_context_target}

For every scene we first fix a base frame sequence $B$, then choose the target
views inside $B$, and finally draw the context views from what remains. For
scenes available at multiple context lengths, the target views remain fixed.

\textbf{DL3DV and RealEstate10K.} $B$ is the native frame order. The targets
are $16$ frames spaced uniformly over $B$. The context pool is $B$ with the
targets removed, order preserved, and $\mathrm{context}(V)$ is its first $V$
frames --- a strict causal prefix of the trajectory.

\textbf{ScanNet.} The 20 ScanNet test sequences are far longer than the other
two benchmarks --- several thousand frames each --- so $B$ is first obtained by
uniformly subsampling every sequence to $1088$ frames ($1024$ context $+$ $64$
targets), which is what makes a $1024$-view context well defined. The targets
are $64$ frames spaced uniformly over $B$.

Each ScanNet context starts with $8$ globally uniform \emph{anchor} frames,
followed by the first $V-8$ context views from the remaining pool. A short pure
prefix covers only the beginning of a room-scale trajectory while targets span
the sequence, so the anchors provide coarse global coverage within the same
budget of $V$ context views. All methods receive identical anchors.

Accordingly, DL3DV and RealEstate10K use $16$ target views per scene, while
ScanNet uses $64$ target views. The indexed context is presented in chunks of
$4$ context views, the targets are never shown to the model, and all methods use
identical index files.

\begin{figure*}[t!]
\centering
\includegraphics[width=\textwidth]{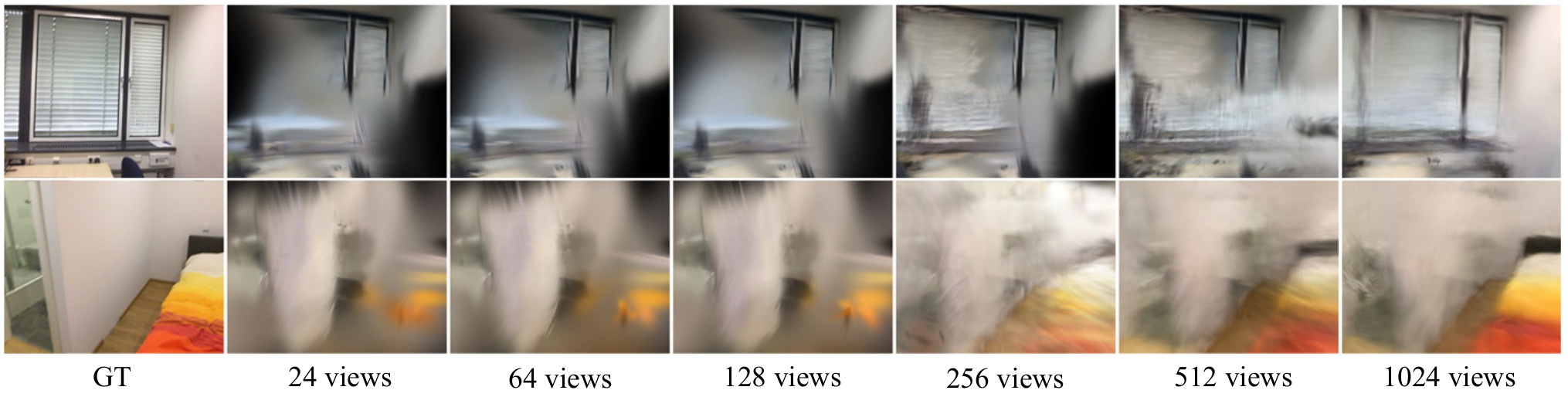}
\caption{Effect of context length on ScanNet. Each row is one scene and shows
the ground-truth target view followed by our renders of that same view, produced
from evaluation contexts containing $24$, $64$, $128$, $256$, $512$ and $1024$
context views. Only the context budget changes along a row.
Regions that no context view has covered yet are reconstructed as low-frequency
blur with shorter contexts and are progressively replaced by observed structure
as the context grows, while content that has already resolved is preserved.}
\label{fig:supp_context}
\end{figure*}

\subsection{Runs, Aggregation and Memory}
\label{sec:supp_runs}

Each reported number corresponds to a \emph{single} evaluation run per
(method, dataset, context length) configuration. The fixed context index removes
evaluation-time sampling variation.
Metrics are first averaged over the target views within a scene and then
averaged over scenes, so every scene contributes equally regardless of how many
targets it provides. Evaluation covers the 140 scenes of the DL3DV-140
benchmark, 140 scenes sampled from the RealEstate10K test split, and the 20
longest ScanNet test scenes. One exception is worth stating: at $V=256$ on
RealEstate10K, $13$ of the $140$ clips are too short to supply $256$ context
views disjoint from the targets, so that column is averaged over the remaining
$127$ scenes; their target views are unchanged and every method evaluated at
$V=256$ uses the same $127$ scenes.

Peak memory is the maximum of the CUDA caching allocator's allocated-bytes
counter over the whole evaluation of a configuration. It therefore covers model weights, activations, the cache and
the rasterizer, but not host memory or the CUDA context itself. Every method is
measured the same way on the same 96\,GB GPU; for baselines we read the same
counter in their own implementation. A configuration is
marked OOM when the run raises a CUDA out-of-memory error before
completing the full scene set, and once a method goes out of memory at some
context length we do not run it at larger ones, since memory grows monotonically
in $V$ for every method considered.

\section{Evaluation Metrics}
\label{sec:supp_metrics}

We report PSNR, SSIM and LPIPS on $[0,1]$ RGB images at the evaluation
resolution of each dataset, with the standard definitions and the standard
reference implementations: SSIM from \texttt{scikit-image} and LPIPS with the
VGG backbone. Naming the implementations matters because SSIM and LPIPS values
are comparable across papers only when the implementations match, and both admit
variants that shift the number appreciably.

The end-to-end latency quoted in the main text is measured with CUDA
synchronization around the timed region: it is the time from receiving a chunk
of $4$ context views to having rendered one novel view from the updated cache,
and excludes data loading, metric computation and writing images to disk.

\section{Additional Experimental Results}
\label{sec:supp_results}

\subsection{Qualitative Effect of Context Length}
\label{sec:supp_context_qualitative}

Tables~\ref{tab:1} and~\ref{tab:2} show that ScanNet accuracy rises from
$13.71$\,dB at $24$ context views to $18.07$\,dB at $1024$ context views.
Figure~\ref{fig:supp_context} visualizes this progression: unobserved regions
appear as low-frequency fill at short contexts and are replaced by structure as
coverage grows, while already-resolved regions remain stable. Residual blur at
$1024$ context views is consistent with the absolute accuracy at that operating
point.

\subsection{Additional Qualitative Comparisons}
\label{sec:supp_more_qualitative}

\begin{figure*}[p]
\centering
\includegraphics[
    width=\textwidth,
    height=0.86\textheight,
    keepaspectratio
]{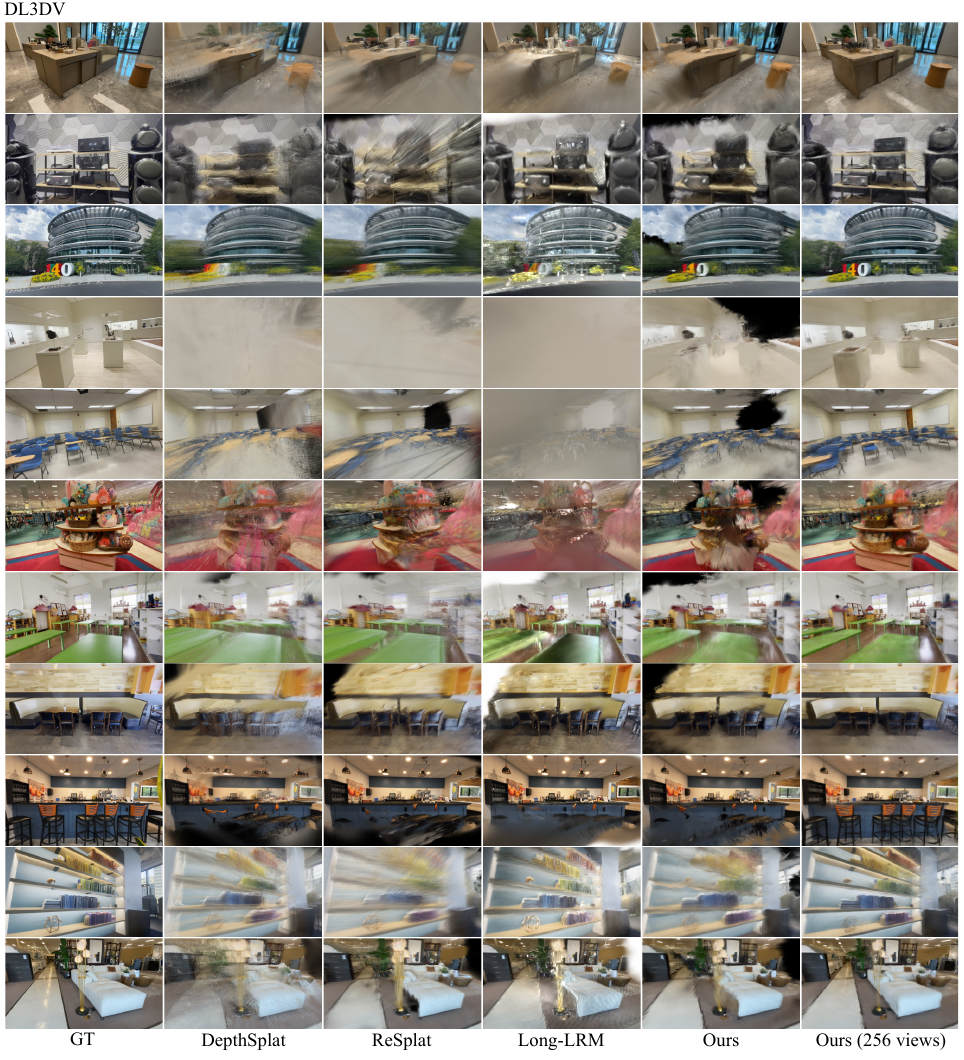}
\caption{Additional qualitative comparisons on DL3DV, in the layout of
Fig.~\ref{fig:qualitative}. DepthSplat, ReSplat, Long-LRM and our directly comparable result use
$64$ context views; the last column shows our reconstruction from $256$ context
views.}
\label{fig:supp_comparison_dl3dv}
\end{figure*}

\begin{figure*}[p]
\centering
\includegraphics[width=\textwidth]
{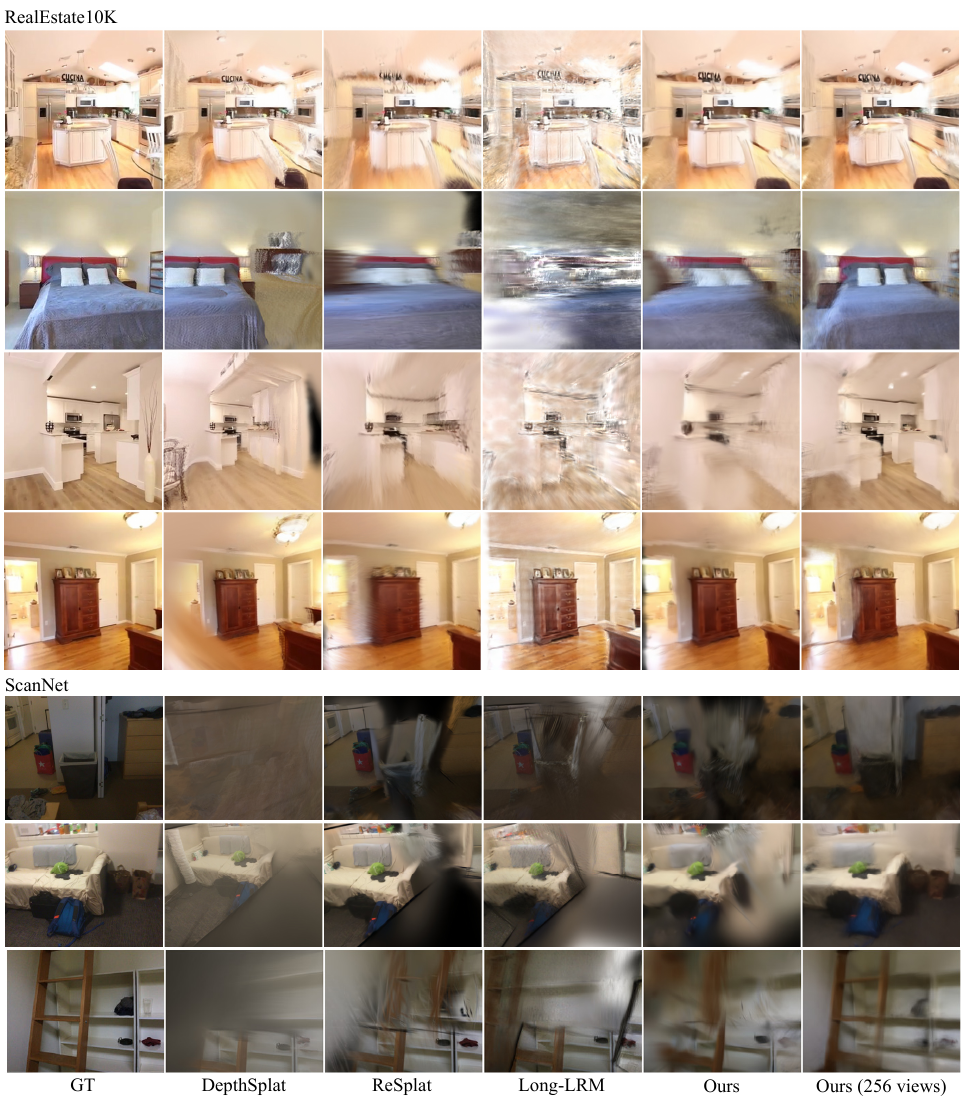}
\caption{Additional qualitative comparisons on RealEstate10K and ScanNet, in
the layout of Fig.~\ref{fig:qualitative}. DepthSplat, ReSplat, Long-LRM and our directly
comparable result use $64$ context views; the last column shows our
reconstruction from $256$ context views.}
\label{fig:supp_comparison_re10k_scannet}
\end{figure*}

Figure~\ref{fig:supp_comparison_dl3dv} adds DL3DV comparisons, and
Figure~\ref{fig:supp_comparison_re10k_scannet} covers RealEstate10K and ScanNet.
All directly comparable results use $64$ context views; the final column shows
our reconstruction from $256$ context views.

\subsection{Additional Feed-Forward Baselines}
\label{sec:supp_extra_baselines}

Table~\ref{tab:supp_extra_baselines} reports pixelSplat
~\cite{charatan2024pixelsplat} and MVSplat~\cite{chen2024mvsplat} at $24$
context views under the same protocol and hardware. MVSplat runs out of memory
at $64$ context views.

\begin{table}[t]
\centering
\caption{Earlier feed-forward baselines evaluated at $24$ context views.
MVSplat runs out of memory at $64$ context views.}
\label{tab:supp_extra_baselines}
\footnotesize
\setlength{\tabcolsep}{3.5pt}
\renewcommand{\arraystretch}{1.08}
\begin{tabular}{llcccc}
\toprule
Method & Dataset & PSNR$\uparrow$ & SSIM$\uparrow$ & LPIPS$\downarrow$ & Mem.$\downarrow$ \\
\midrule
\multirow{3}{*}{MVSplat}
 & DL3DV   & 11.51 & 0.274 & 0.611 & 23.4 \\
 & RealEstate10K & 12.01 & 0.417 & 0.514 & 23.3 \\
 & ScanNet &  9.46 & 0.200 & 0.685 & 37.7 \\
\midrule
\multirow{3}{*}{pixelSplat}
 & DL3DV   & \multicolumn{4}{c}{\multirow{3}{*}{OOM}} \\
 & RealEstate10K & \multicolumn{4}{c}{} \\
 & ScanNet & \multicolumn{4}{c}{} \\
\bottomrule
\end{tabular}
\end{table}

\subsection{Ablation Details}
\label{sec:supp_ablation}

Variant definitions follow the ablation study in the main text, and all
variants use the protocol in Sec.~\ref{sec:supp_protocol}. Every ablated variant
is retrained with its stated configuration; results are not obtained by
disabling modules at inference time in a shared full-model checkpoint.

\section{Limitations and Future Work}
\label{sec:supp_limitations}

\emph{StreamSplat} assumes calibrated context views with known camera parameters,
so its reconstruction quality depends on the accuracy of the provided poses;
relaxing this assumption toward pose-free streaming reconstruction is an
important next step. Our formulation also targets static scenes and does not
explicitly model moving objects or lighting changes that arise in real video
streams. Finally, the voxel cache uses a fixed voxel size and pivot budget,
which couples memory usage with geometric detail; adaptive voxel resolution and
content-aware cache allocation could further improve the accuracy--memory
trade-off at long context lengths.

A further limitation is visible in the results above: on short, densely-sampled
trajectories such as RealEstate10K clips, offline methods that attend over all
context views jointly remain more accurate than causal processing at the context
lengths where they still fit in memory. Closing that gap without giving up the
streaming property --- for example by allowing a bounded amount of retrospective
revision of cached geometry --- is the most direct avenue for improving the
method.


\begin{thebibliography}{30}
\providecommand{\natexlab}[1]{#1}

\bibitem[{Charatan et~al.(2024)Charatan, Li, Tagliasacchi, and
  Sitzmann}]{charatan2024pixelsplat}
Charatan, D.; Li, S.~L.; Tagliasacchi, A.; and Sitzmann, V. 2024.
\newblock pixelSplat: 3D Gaussian Splats from Image Pairs for Scalable
  Generalizable 3D Reconstruction.
\newblock In \emph{Proceedings of the IEEE/CVF Conference on Computer Vision
  and Pattern Recognition}, 19457--19467.

\bibitem[{Chen et~al.(2026)Chen, Li, Zhou, Li, Ma, and Guo}]{chen2026of3gs}
Chen, R.; Li, F.; Zhou, C.; Li, Z.; Ma, Z.; and Guo, H. 2026.
\newblock {OF$^3$GS}: On-the-Fly Feed-Forward 3D Gaussian Splatting from
  Unposed Images.
\newblock arXiv:2606.03254.

\bibitem[{Chen et~al.(2024{\natexlab{a}})Chen, Xu, Zheng, Zhuang, Pollefeys,
  Geiger, Cham, and Cai}]{chen2024mvsplat}
Chen, Y.; Xu, H.; Zheng, C.; Zhuang, B.; Pollefeys, M.; Geiger, A.; Cham,
  T.-J.; and Cai, J. 2024{\natexlab{a}}.
\newblock {MVSplat}: Efficient 3D Gaussian Splatting from Sparse Multi-View
  Images.
\newblock In \emph{European Conference on Computer Vision}.

\bibitem[{Chen et~al.(2024{\natexlab{b}})Chen, Tan, Zhang, Bi, Luan, Hong, Li,
  and Xu}]{chen2024longlrm}
Chen, Z.; Tan, H.; Zhang, K.; Bi, S.; Luan, F.; Hong, Y.; Li, F.; and Xu, Z.
  2024{\natexlab{b}}.
\newblock {Long-LRM}: Long-Sequence Large Reconstruction Model for
  Wide-Coverage Gaussian Splats.
\newblock \emph{arXiv preprint arXiv:2410.12781}.

\bibitem[{Cheng et~al.(2026)Cheng, Ye, Li, You, Zhan, and
  Yang}]{cheng2026recosplat}
Cheng, F.; Ye, B.; Li, X.; You, J.; Zhan, F.; and Yang, M.-H. 2026.
\newblock ReCoSplat: Autoregressive Feed-Forward Gaussian Splatting Using
  Render-and-Compare.
\newblock arXiv:2603.09968.

\bibitem[{Dai et~al.(2017)Dai, Chang, Savva, Halber, Funkhouser, and
  Nie{\ss}ner}]{dai2017scannet}
Dai, A.; Chang, A.~X.; Savva, M.; Halber, M.; Funkhouser, T.; and Nie{\ss}ner,
  M. 2017.
\newblock {ScanNet}: Richly-Annotated 3D Reconstructions of Indoor Scenes.
\newblock In \emph{Proceedings of the IEEE Conference on Computer Vision and
  Pattern Recognition}, 5828--5839.

\bibitem[{Fang and Wang(2024)}]{fang2024mini}
Fang, G.; and Wang, B. 2024.
\newblock Mini-Splatting: Representing Scenes with a Constrained Number of
  Gaussians.
\newblock In \emph{European Conference on Computer Vision}, 165--181.

\bibitem[{Hahlbohm et~al.(2026)Hahlbohm, Franke, Eisemann, and
  Magnor}]{hahlbohm2026faster}
Hahlbohm, F.; Franke, L.; Eisemann, M.; and Magnor, M. 2026.
\newblock Faster-GS: Analyzing and Improving Gaussian Splatting Optimization.
\newblock In \emph{Proceedings of the IEEE/CVF Conference on Computer Vision
  and Pattern Recognition}, 18946--18957.

\bibitem[{Huang et~al.(2026)Huang, Wang, Gao, Sun, Wu, Gao, and
  Jia}]{huang2026longsplat}
Huang, G.; Wang, R.; Gao, X.; Sun, C.; Wu, Y.; Gao, S.; and Jia, Y. 2026.
\newblock {LongSplat}: Online Generalizable 3D Gaussian Splatting from Long
  Sequence Images.
\newblock In \emph{Proceedings of the AAAI Conference on Artificial
  Intelligence}, volume~40, 4994--5002.

\bibitem[{Jiang et~al.(2025)Jiang, Mao, Xu, Lu, Ren, Jin, Xu, Yu, Pang, Zhao,
  Lin, and Dai}]{jiang2025anysplat}
Jiang, L.; Mao, Y.; Xu, L.; Lu, T.; Ren, K.; Jin, Y.; Xu, X.; Yu, M.; Pang, J.;
  Zhao, F.; Lin, D.; and Dai, B. 2025.
\newblock {AnySplat}: Feed-Forward 3D Gaussian Splatting from Unconstrained
  Views.
\newblock \emph{ACM Transactions on Graphics}, 44(6).

\bibitem[{Kerbl et~al.(2023)Kerbl, Kopanas, Leimk{\"u}hler, and
  Drettakis}]{kerbl2023gaussians}
Kerbl, B.; Kopanas, G.; Leimk{\"u}hler, T.; and Drettakis, G. 2023.
\newblock 3D Gaussian Splatting for Real-Time Radiance Field Rendering.
\newblock \emph{ACM Transactions on Graphics}, 42(4).

\bibitem[{Kotovenko, Grebenkova, and Ommer(2026)}]{kotovenko2026edgs}
Kotovenko, D.; Grebenkova, O.; and Ommer, B. 2026.
\newblock {EDGS}: Eliminating Densification for Efficient Convergence of 3DGS.
\newblock In \emph{Proceedings of the IEEE/CVF Conference on Computer Vision
  and Pattern Recognition}, 41065--41076.

\bibitem[{Li et~al.(2026)Li, Lv, Tang, Yang, and Huang}]{li2026tokensplat}
Li, Y.; Lv, C.; Tang, Z.; Yang, H.; and Huang, D. 2026.
\newblock {TokenSplat}: Token-aligned 3D Gaussian Splatting for Feed-forward
  Pose-free Reconstruction.
\newblock In \emph{Proceedings of the IEEE/CVF Conference on Computer Vision
  and Pattern Recognition}, 40886--40895.

\bibitem[{Li et~al.(2025)Li, Wang, Chu, Li, Kao, Chen, and Lu}]{li2025streamgs}
Li, Y.; Wang, J.; Chu, L.; Li, X.; Kao, S.-H.; Chen, Y.-C.; and Lu, Y. 2025.
\newblock {StreamGS}: Online Generalizable Gaussian Splatting Reconstruction
  for Unposed Image Streams.
\newblock In \emph{Proceedings of the IEEE/CVF International Conference on
  Computer Vision}, 25841--25850.

\bibitem[{Ling et~al.(2024)Ling, Sheng, Tu, Zhao, Xin, Wan, Yu, Guo, Yu, Lu,
  Li, Sun, Ashok, Mukherjee, Kang, Kong, Hua, Zhang, Benes, and
  Bera}]{ling2024dl3dv}
Ling, L.; Sheng, Y.; Tu, Z.; Zhao, W.; Xin, C.; Wan, K.; Yu, L.; Guo, Q.; Yu,
  Z.; Lu, Y.; Li, X.; Sun, X.; Ashok, R.; Mukherjee, A.; Kang, H.; Kong, X.;
  Hua, G.; Zhang, T.; Benes, B.; and Bera, A. 2024.
\newblock {DL3DV-10K}: A Large-Scale Scene Dataset for Deep Learning-based 3D
  Vision.
\newblock In \emph{Proceedings of the IEEE/CVF Conference on Computer Vision
  and Pattern Recognition}.

\bibitem[{Mildenhall et~al.(2021)Mildenhall, Srinivasan, Tancik, Barron,
  Ramamoorthi, and Ng}]{mildenhall2021nerf}
Mildenhall, B.; Srinivasan, P.~P.; Tancik, M.; Barron, J.~T.; Ramamoorthi, R.;
  and Ng, R. 2021.
\newblock {NeRF}: Representing Scenes as Neural Radiance Fields for View
  Synthesis.
\newblock \emph{Communications of the ACM}, 65(1): 99--106.

\bibitem[{Mo, Cai, and Liu(2026)}]{Mo_2026_CVPR}
Mo, Y.; Cai, Y.; and Liu, L. 2026.
\newblock Plug-and-Play PDE Optimization for 3D Gaussian Splatting: Toward
  High-Quality Rendering and Reconstruction.
\newblock In \emph{Proceedings of the IEEE/CVF Conference on Computer Vision
  and Pattern Recognition}, 33333--33342.

\bibitem[{Ren et~al.(2026)Ren, Wen, Fang, and Lu}]{ren2026fastgs}
Ren, S.; Wen, T.; Fang, Y.; and Lu, B. 2026.
\newblock {FastGS}: Training 3D Gaussian Splatting in 100 Seconds.
\newblock In \emph{Proceedings of the IEEE/CVF Conference on Computer Vision
  and Pattern Recognition}, 26094--26103.

\bibitem[{Smart et~al.(2024)Smart, Zheng, Laina, and
  Prisacariu}]{smart2024splatt3r}
Smart, B.; Zheng, C.; Laina, I.; and Prisacariu, V.~A. 2024.
\newblock {Splatt3R}: Zero-Shot Gaussian Splatting from Uncalibrated Image
  Pairs.
\newblock \emph{arXiv preprint arXiv:2408.13912}.

\bibitem[{Veicht et~al.(2026)Veicht, Hong, Barath, and
  Pollefeys}]{veicht2026zipsplat}
Veicht, A.; Hong, S.; Barath, D.; and Pollefeys, M. 2026.
\newblock {ZipSplat}: Fewer Gaussians, Better Splats.
\newblock \emph{arXiv preprint arXiv:2606.05102}.

\bibitem[{Wang et~al.(2026)Wang, Song, Cai, and Liu}]{wang2026stac}
Wang, R.; Song, Y.; Cai, Y.; and Liu, L. 2026.
\newblock {STAC}: Plug-and-Play Spatio-Temporal Aware Cache Compression for
  Streaming 3D Reconstruction.
\newblock In \emph{Proceedings of the IEEE/CVF Conference on Computer Vision
  and Pattern Recognition}, 7567--7576.

\bibitem[{Wang et~al.(2004)Wang, Bovik, Sheikh, and Simoncelli}]{wang2004image}
Wang, Z.; Bovik, A.~C.; Sheikh, H.~R.; and Simoncelli, E.~P. 2004.
\newblock Image Quality Assessment: From Error Visibility to Structural
  Similarity.
\newblock \emph{IEEE Transactions on Image Processing}, 13(4): 600--612.

\bibitem[{Wu et~al.(2026)Wu, Yan, Yi, Wang, and Liao}]{wu2026streamsplat}
Wu, Z.; Yan, Q.; Yi, X.; Wang, L.; and Liao, R. 2026.
\newblock {StreamSplat}: Towards Online Dynamic 3D Reconstruction from
  Uncalibrated Video Streams.
\newblock In \emph{International Conference on Learning Representations}.

\bibitem[{Xu et~al.(2025{\natexlab{a}})Xu, Barath, Geiger, and
  Pollefeys}]{xu2025resplat}
Xu, H.; Barath, D.; Geiger, A.; and Pollefeys, M. 2025{\natexlab{a}}.
\newblock {ReSplat}: Learning Recurrent Gaussian Splats.
\newblock \emph{arXiv preprint arXiv:2510.08575}.

\bibitem[{Xu et~al.(2025{\natexlab{b}})Xu, Peng, Wang, Blum, Barath, Geiger,
  and Pollefeys}]{xu2025depthsplat}
Xu, H.; Peng, S.; Wang, F.; Blum, H.; Barath, D.; Geiger, A.; and Pollefeys, M.
  2025{\natexlab{b}}.
\newblock {DepthSplat}: Connecting Gaussian Splatting and Depth.
\newblock In \emph{Proceedings of the IEEE/CVF Conference on Computer Vision
  and Pattern Recognition}, 16453--16463.

\bibitem[{Ye et~al.(2025)Ye, Liu, Xu, Li, Pollefeys, Yang, and
  Peng}]{ye2025noposplat}
Ye, B.; Liu, S.; Xu, H.; Li, X.; Pollefeys, M.; Yang, M.-H.; and Peng, S. 2025.
\newblock No Pose, No Problem: Surprisingly Simple 3D Gaussian Splats from
  Sparse Unposed Images.
\newblock In \emph{International Conference on Learning Representations}.

\bibitem[{Yu et~al.(2024)Yu, Chen, Huang, Sattler, and Geiger}]{yu2024mip}
Yu, Z.; Chen, A.; Huang, B.; Sattler, T.; and Geiger, A. 2024.
\newblock Mip-Splatting: Alias-Free 3D Gaussian Splatting.
\newblock In \emph{Proceedings of the IEEE/CVF Conference on Computer Vision
  and Pattern Recognition}, 19447--19456.

\bibitem[{Zhang et~al.(2018)Zhang, Isola, Efros, Shechtman, and
  Wang}]{zhang2018unreasonable}
Zhang, R.; Isola, P.; Efros, A.~A.; Shechtman, E.; and Wang, O. 2018.
\newblock The Unreasonable Effectiveness of Deep Features as a Perceptual
  Metric.
\newblock In \emph{Proceedings of the IEEE Conference on Computer Vision and
  Pattern Recognition}, 586--595.

\bibitem[{Zhou et~al.(2018)Zhou, Tucker, Flynn, Fyffe, and
  Snavely}]{zhou2018stereo}
Zhou, T.; Tucker, R.; Flynn, J.; Fyffe, G.; and Snavely, N. 2018.
\newblock Stereo Magnification: Learning View Synthesis Using Multiplane
  Images.
\newblock \emph{ACM Transactions on Graphics}, 37(4): 65:1--65:12.

\bibitem[{Zhuo et~al.(2025)Zhuo, Chen, Liao, and Hu}]{zhuo2025streaming4d}
Zhuo, L.; Chen, Y.; Liao, S.; and Hu, H. 2025.
\newblock Streaming 4D Visual Geometry Transformer.
\newblock \emph{arXiv preprint arXiv:2507.11539}.

\end{thebibliography}
\end{document}